\documentclass[10pt, conference, compsocconf]{IEEEtran}

\usepackage{amsmath,amssymb,amsfonts}
\usepackage{caption}
\usepackage{algorithmic}
\usepackage{array}
\usepackage{graphicx}
\usepackage{textcomp}
\usepackage{xcolor}
\usepackage{url}
\usepackage{siunitx}
\usepackage{pgfplots}
\pgfplotsset{compat=newest}
\usepackage{booktabs}
\usepackage{multirow}
\usepackage{float}
\usepackage{relsize}
\floatstyle{plaintop}
\restylefloat{table}
\def\BibTeX{{\rm B\kern-.05em{\sc i\kern-.025em b}\kern-.08em
		T\kern-.1667em\lower.7ex\hbox{E}\kern-.125emX}}

\begin{document}
	\title{Real-Time Optimized \textit{N}-gram For Mobile Devices}

\author{\IEEEauthorblockN{
		Sharmila Mani,
		Sourabh Vasant Gothe,
		Sourav Ghosh,
		Ajay Kumar Mishra,\\
		Prakhar Kulshreshtha,
		Bhargavi M and 
		Muthu Kumaran 
	}
	\IEEEauthorblockA{Samsung R\&D Institute Bangalore,
		Karnataka, India 560037\\
		Email: \{sharmila.m,  sourab.gothe, sourav.ghosh, ajay.mishra, p.kulshresht, bhargavi.m, muthuk\}@samsung.com}
	}

\makeatletter
\def\ps@IEEEtitlepagestyle{
  \def\@oddfoot{\mycopyrightnotice}
  \def\@evenfoot{}
}
\def\mycopyrightnotice{
  {\footnotesize
  \begin{minipage}{\textwidth}
  \centering
  \hrule\vspace{6pt}
  978-1-5386-6783-5/19/\$31.00~\copyright~2019 IEEE.
  Personal use of this material is permitted.  Permission from IEEE must be obtained for all other uses, in any current or future media, including reprinting/republishing this material for advertising or promotional purposes, creating new collective works, for resale or redistribution to servers or lists, or reuse of any copyrighted component of this work in other works.\\
  https://doi.org/10.1109/ICOSC.2019.8665639
  \end{minipage}
  }
}

\maketitle

\begin{abstract}With the increasing number of mobile devices, there has been continuous research on generating optimized Language Models (LMs) for soft keyboard. In spite of advances in this domain, building a single LM for low-end feature phones as well as high-end smartphones is still a pressing need. Hence, we propose a novel technique, Optimized \textit{N}-gram (Op-Ngram), an end-to-end \textit{N}-gram pipeline that utilises mobile resources efficiently for faster Word Completion (WC) and Next Word Prediction (NWP). Op-Ngram applies Stupid Backoff \cite{brants2007large} and pruning strategies to generate a light-weight model. The LM loading time on mobile is linear with respect to model size. We observed that Op-Ngram gives 37\% improvement in Language Model (LM)-ROM size, 76\% in LM-RAM size, 88\% in loading time and 89\% in average suggestion time as compared to \textsc{Sorted} array variant of BerkeleyLM\cite{pauls2011faster}. Moreover, our method shows significant performance improvement over KenLM\cite{heafield2011kenlm} as well.
\end{abstract}
	\begin{IEEEkeywords} Language Modelling, Mobile Devices, Natural Language Processing, \textit{N}-gram, Soft Keyboard.\end{IEEEkeywords}

	\section{Introduction}
	Language modelling is one of the crucial aspects of Natural Language Processing. It involves representing statistical language information to estimate a probability distribution over a sequence of words. $N$-gram models \cite{doi:10.1002/j.1538-7305.1948.tb01338.x}, which are Markov chain based (or count-based) LMs to predict the probability of $n^{\text{th}}$ word based on the preceding sequence of $(n-1)$ words, have been widely used for language modelling tasks. 
	
\begin{figure}
		\centering
		\includegraphics[width=0.8\linewidth]{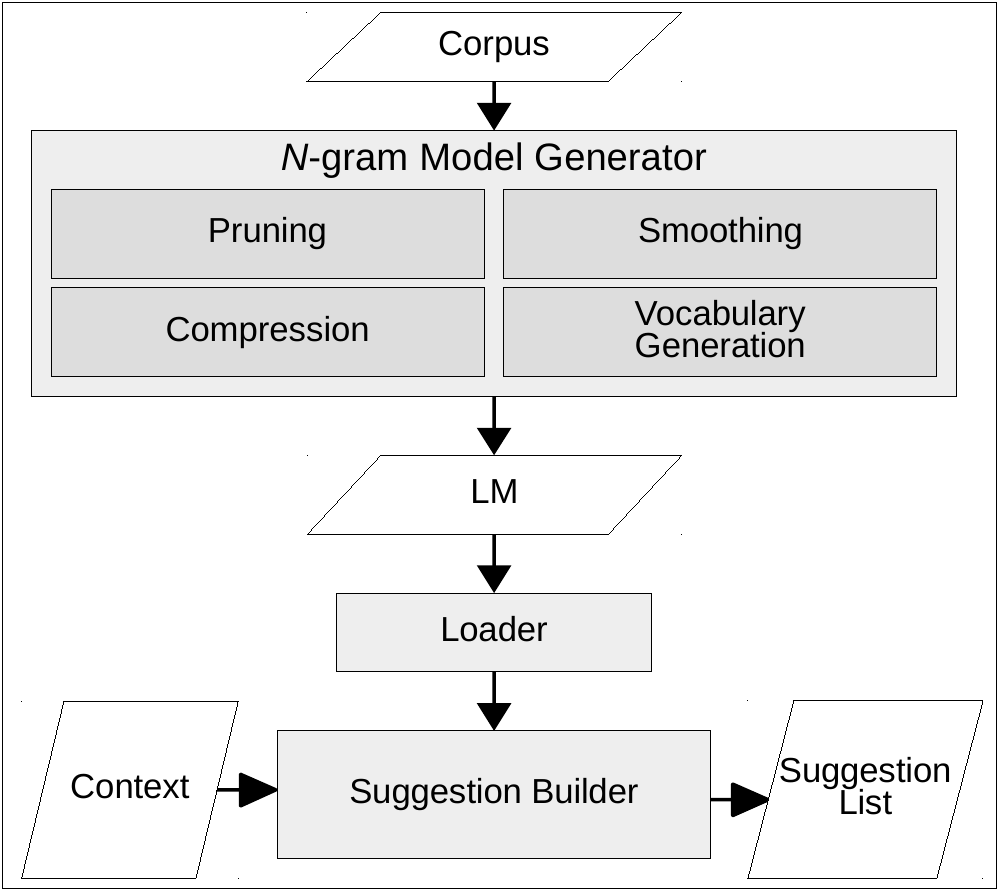}
		\caption{Op-Ngram Flow}
		\label{fig:tymeNgramArchitecture}
	\end{figure}

	One of the important applications of language model is text suggestion in soft keyboards. The challenges involved in language modelling for mobile devices not only include providing better text suggestion but also providing it in real-time. The real-time model must be lightweight and the inference engine be fast enough. In this context, we present in this paper Optimized $N$-gram (Op-Ngram) that describes an end-to-end $N$-gram pipeline (Figure \ref{fig:tymeNgramArchitecture}). Op-Ngram involves optimisations from (a) generating a light-weight LM with efficient structure,  to (b) inferencing of the model using a for seamless Word Completions and Next Word Predictions. 
	
	The optimisations in Op-Ngram for achieving light-weight model include Stupid Backoff \cite{brants2007large} and perplexity-based pruning. In $N$-gram models, backoff is used to normalise probabilities by penalising use of lower-order $N$-grams. By using Stupid Backoff, backoff values need not be pre-computed and stored for every combination, but a single heuristic constant can be used as the common backoff factor. For text suggestions in mobile devices, the goal is to provide top-$K$ suggestions, where $K < 10$. Thus, less frequent words can be skipped from the generated model. These factors help in reducing our model size. Additionally, fast model loading and real-time inference on mobile has been achieved using a unique storage strategy. The model has been distributed across multiple files to enable parallel model loading.
	
	In the next section, we discuss relevant $N$-gram approaches for text suggestions available in literature. In section \ref{sec:proposedModel}, we present Op-Ngram in detail, which includes model generation, model storage, and model inference. In section \ref{sec:experimentResults}, we also present results of Op-Ngram with English and Hindi in comparison to the available literature.

	\section{Related Work}\label{sec:relatedWork}
	Paul et al. \cite{pauls2011faster} describes a number of fast and compact approaches to language modelling. One simple variant discussed uses \textsc{Sorted} arrays for storing the LM, which we refer to as \textsc{Sorted} BerkeleyLM (SBLM). For each order of $N$-gram, this model encodes $\left(w_n, c\left(w_1^{n-1}\right)\right)$ using \textit{context encoding} and uses these values as keys in a map. Here, $c$ refers to entries in a $\left(n-1\right)$-gram sorted array. $N$-gram information is stored in blocks containing probability, backoff and index values. The indices of $\left(n-1\right)$-gram, or $\left(n-1\right)^\text{th}$ order $N$-gram, is used to identify terms in $n^\text{th}$ order. Lookup by $N$-grams in this approach is logarithmic as binary search is used on the sorted array. Although this LM is faster than many other models proposed in literature, but the model size is not efficient in terms of real-time usage on a mobile device.
	
	KenLM \cite{heafield2011kenlm} is another widely used $N$-gram LM which introduces optimisations to model size using a sorted array and \textsc{Trie} data structure, and optimisations to speed by using hash tables and linear \textsc{Probing}. Like SBLM, the KenLM \textsc{Trie} approach uses reverse trie and separate  suffix-sorted arrays for each order of $N$-gram, and in \textsc{Probing}, number of hash table buckets should be more than that of entries to ensure collision resolution. The approaches discussed in this work optimises either speed or memory over the other, but not both at once.
	
	In the following sections, we discuss our approach which is optimized for speed and memory without compromising one for the other. This manifests in terms of loading time, average suggestion time, RAM and ROM size with little impact on the relevance of text suggestions.

	\section{Proposed Model (Op-Ngram)}\label{sec:proposedModel}
	$N$-gram is a statistical model which gives the probability distribution over a vocabulary $V$ given the previous $N-1$ words as context. The probability assigned to a word sequence is:
	
	\begin{equation}\label{eqn:probWordSequence}
		\small
		P\left(w_n\middle|w_{n-N+1}^{n-1}\right) =
		\begin{cases}
			\frac{\left(1-d\right)\left\lvert w_{n-N+1}^n \right\rvert}{\left\lvert w_{n-N+1}^{n-1} \right\rvert}, & \left\lvert w_{n-N+1}^n \right\rvert > 0 \\
		\lambda\left( w_{n-N+1}^{n-1}\right) \times\\  P\left(w_n\middle|w_{n-N+2}^{n-1}\right), & \left\lvert w_{n-N+1}^n\right\rvert = 0
		\end{cases}
	\end{equation}
	
	where $w_j^k$ represents the word sequence ($w_j$,$w_{j+1}$,...,$w_k$), $d$ refers to a discount ratio, $\left\lvert w_1^n \right\rvert$ indicates the count of $w_1^n$, and the function $\lambda$ represents the back-off weight to ensure that probability values add up to unity \cite{brants2007large}.
	
	We use trigram model $\left(N=3\right)$ as trigrams are known to be complex-enough for effective predictions, and simple enough for storing and retrieving them easily.	Figure \ref{fig:tymeNgramArchitecture} details the flow of Op-Ngram.

	\subsection{Model Generation}
	This subsection describes the model generation pipeline. The pre-processing phase involves cleaning and sampling a crawled corpus, fetching word-count statistics, marking sentence-endings, and tagging blacklisted and unknown words in the corpus text, where we leverage Hadoop-based distributed processing. A sampling size of $2.5$ GB is considered to have an optimal trade-off between model inference and size. The beginning and ending of a sentence is marked by prefixing a start-tag, \texttt{<s>}, and an end-tag, \texttt{<e>}, to preserve the probability of words appearing at the starting and end of a sentence. A linguist-verified bad words list is used to filter out common blacklisted words and replace their occurrences with \texttt{<bad>} tag. Any word that occurs in the corpus text with a frequency below a pre-defined threshold is deemed to be a rare or unknown word and is substituted with \texttt{<unk>} tag. Tagging of blacklisted and rare words helps to minimise their impact on the model quality.
	
	During model generation, the counts of unigrams, bigrams, and trigrams are obtained from the pre-processed corpus. However, if all the $N$-grams are to be included then model size becomes comparable to original corpus size (i.e. $\sim2$ GB). So, non-informative bigrams and trigrams are pruned for reducing the model size. Brants et al. \cite{brants2007large} proposes that at a very high pruning threshold even simple count-based pruning gives similar results to complicated methods like Stolke-Pruning \cite{DBLP:journals/corr/cs-CL-0006025}. Op-Ngram uses perplexity-based pruning, where the importance of a trigram (for pruning) is defined as:
	
	\begin{equation}\label{eqn:perplexityBasedPruning}
		\small
		S\left(w_1^3\right) = \left\lvert w_1^3\right\rvert \cdot \left(P\left(w_3\middle|w_1^2\right) - \alpha\left(\left\lvert w_1^2\right\rvert \right) \cdot P\left(w_2\middle|w_1\right)\right)
	\end{equation}
	
\begin{figure}
		\centering
		\pgfplotsset{
			width=\linewidth
		}
		\begin{tikzpicture}
		\begin{axis}[xlabel=Vocabulary Size ($\times 100k$ words),ylabel=Corpus Coverage (\%),
		xtick = {0,0.2,0.4,0.6,0.8,1,1.2,1.4,1.6},
		xticklabel style={/pgf/number format/fixed}
		]
		\addplot[smooth,mark=*] coordinates {
			(0.01, 62)
			(0.05, 79)
			(0.1, 84)
			(0.25, 87.5)
			(0.5, 90)
			(1, 91)
			(1.5, 91.2)
		};
		\addplot[dashed] coordinates{
			(1, 60) (1, 100)
		};
		\end{axis}
		\end{tikzpicture}
		\caption{Corpus Coverage over varying Vocabulary Size}
		\label{fig:vocabSizeComparison}
	\end{figure}
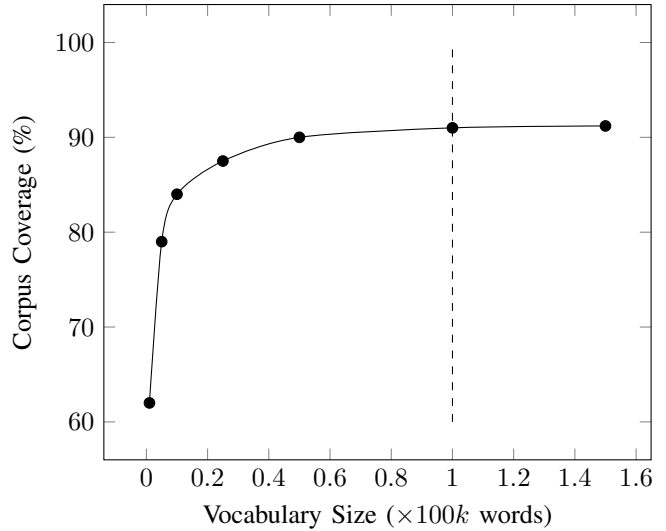

	Based on experimental results given in Figure \ref{fig:vocabSizeComparison}, only the top-$100k$ unigrams in terms of frequency are considered and denoted as $n_{uni}$ that is equal to vocabulary size $V$. Maximum allowed number of bigrams and trigrams, $n_{bi}$ and $n_{tri}$ respectively, are set at $200k$ and $250k$ on the basis of similar experiments. It may be noted that no bigram refers a unigram not present in the top-$100k$ list and no trigram refers a context bigram not present in the top-$200k$ bigram list.
	
\begin{figure}
		\centering
		\includegraphics[width=\linewidth]{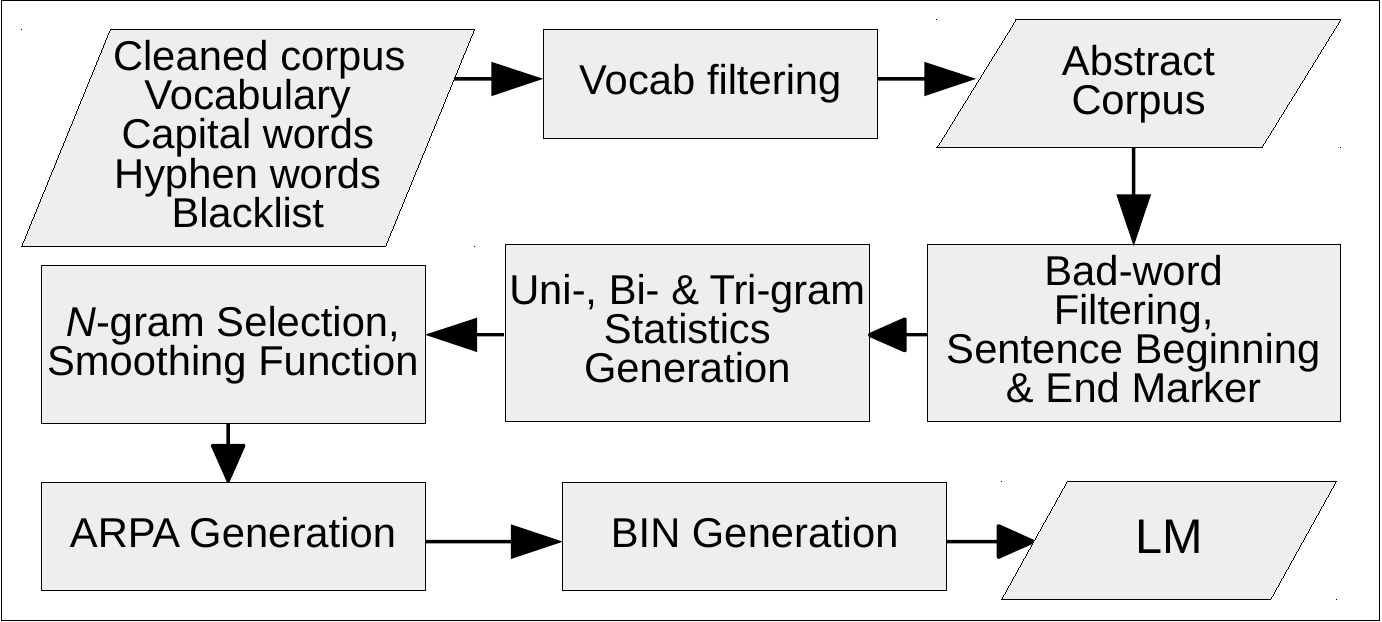}
		\caption{$N$-gram model generation pipeline}
		\label{fig:ngramModelGenerationPipeline}
	\end{figure}

	To assign backoff values, we experimented with different smoothing techniques like: (a) Absolute Discounting, (b) Kneser-Ney smoothing \cite{kneser1995improved}, (c) Stupid Backoff. As demonstrated in literature \cite{brants2007large}, we found that Stupid Backoff method, in spite of being relatively simpler, works at least as well as most other well-known smoothing techniques in case of highly pruned models like ours. The selected $N$-grams, along with their corresponding probability values, are written out to an \texttt{ARPA} file,  that stores a  statistical $N$-gram model. This \texttt{ARPA} file is processed to create two binary files as explained in the next subsection. An overview of the $N$-gram model generation pipeline is given in Figure \ref{fig:ngramModelGenerationPipeline}.
	
	When the context is unavailable in $N$-gram model, we incorporate the Class-$N$-gram model to improve the relevance of suggestions. This complements the $N$-gram model by identifying the sequence of classes which yields the previous words, and inferring the maximum likelihood estimate class(es) for the next word in the sequence. Unlike an unsupervised clustering of words, we derive a word-to-class mapping based on part-of-speech (POS) tagging of training text. To make the model language independent, we have selected a POS tagger that ignores language-specific features like capitalisation \cite{reddy2011cross}. This feature-independence is also crucial for our purpose as users do not always provide correct capitalisation during casual typing in a soft keyboard.
	Bharati et al. \cite{bharati2006anncorra} proposes POS tagger design that unifies word variations based on attributes like gender, time, etc. This effectively reduces the number of classes with little negative impact on model inference, and thereby, makes inference computation faster.
	
	As discussed in literature \cite{pantel2002discovering}, in many languages, one word may appear in multiple contexts, senses and/or parts-of-speech. However, for our implementation, we have taken into account, only the single most probable POS tag that applies to each word to maintain low model size.	
	
	\subsection{Model Storage}\label{sec:modelStorage}
	
	\subsubsection{$N$-gram model} 
\begin{figure}
		\centering
		\includegraphics[width=\linewidth]{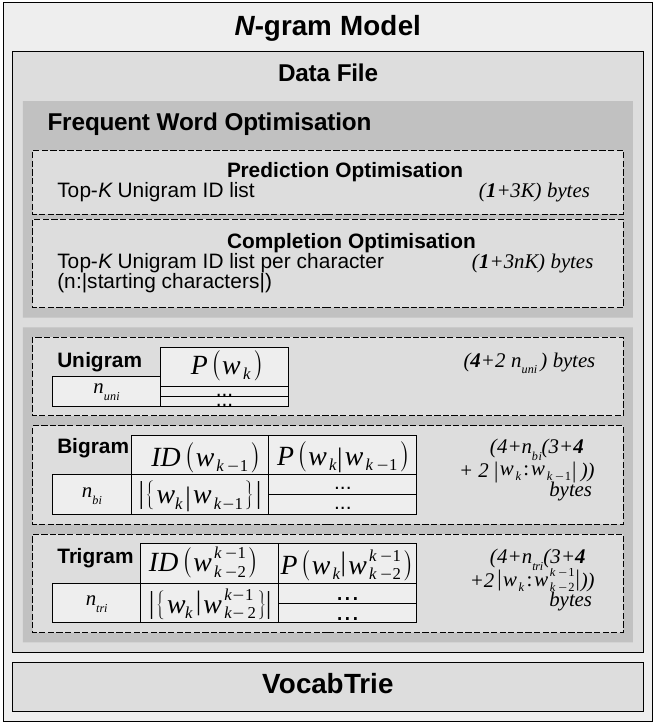}
		\caption{Storage strategy and Bytes required for $N$-gram model of Op-Ngram
			\textit{(Numbers in bold indicate bytes consumed for storing counts.
				$\left\lvert \left\{ w_k \middle| w_x^y \right\} \right\rvert$ indicates the number of unigrams, $w_k$, for each context $N$-gram, $w_x^y$.
				Refer Section \ref{sec:proposedModel} for other notations.)
			}
		}
		\label{fig:tymeNgramStorageStrategy}
	\end{figure}

The $N$-gram model is saved as two binary files: the unigram vocabulary in a VocabTrie \cite{marisa}, and word IDs with probabilities in another file, as depicted in Figure \ref{fig:tymeNgramStorageStrategy}. We have introduced a number of optimisations in the storage strategy. One of them is storing only the unigram probability values, each of which take just 2 bytes. The unigram ID is implicit based on the order in which they appear in the data file, starting from $0$. We have achieved this 2-byte upper bound by storing $min\left(\left\lfloor-10^{c_1} \cdot \log p_i\right\rfloor, c_2\right)$ instead of $p_i$ which represents the probability values corresponding to the $i^{\text{th}}$ unigram. The exponent $c_1$ is set to $3$ and the maximum allowed value $c_2$ is fixed at $29999$, thus enabling us to store each value within $2$ bytes with negligible loss in floating-point precision.
	
	Optimisations for real-time suggestion include Frequent Word Optimisation (FWO), with (a) Prediction Optimisation: top-$K$ unigram IDs for faster lookup in the absence of context,  ex. \textit{\{ ``the'', ``of'', .. \}}, and (b) Completion Optimisation: list of top-$K$ unigram IDs per character mapped with its respective starting alphabet in the language, ex. \textit{[`a' : \{ ``and'', ``ant'', .. \}, `b' : \{ ``bag'', ``ball'', .. \}]}. The former helps in faster Next Word Prediction and the latter prevents looking up the entire language model during Word Completion when single character prefix is provided. This improves the average suggestion building time by a significant amount.
	
	The data file is compressed using zlib compression \cite{zlib} to reduce the ROM size 
	at the cost of negligible addition to loading time on account of decompression overhead.
	
	\subsubsection{Class-$N$-gram model}The Class-$N$-gram model for Op-Ngram is stored in a single binary file containing three sections:

	(a) word-to-class mapping, (b) top-$K$ unigrams per class, (c) most likely classes for all context pairs. The first section, the word-to-class mapping, is in the form of an array of integers representing class IDs that consume $1$-byte each. The unigram IDs from Vocab Trie are used as the indices for this array which results in $O(1)$ lookup of class IDs. Although using $1$-byte per class ID enables the use of $256$ unique classes, Op-Ngram uses ($\sim 32$) classes. In contrast, a large number of classes does not provide additional benefit, but leads to increase in the model size. The next section, list of top-$K$ unigrams per class is a sequence of unigram IDs mapped to a class, each unigram ID consuming $3$-bytes as in $N$-gram data file. These top-$K$ words are selected based on the highest values of conditional probability of word $w_k$ given class $C_k$, $P\left(w_k\middle|C_k\right)$. The last section, most likely class information, stores only a sequence of the IDs of highest probable classes $C_k$ corresponding to a sorted sequence of all context class pairs. Thus, for $32$ classes, there are $1024 \left(=32^2\right)$ IDs stored in this section. In summary, Class-$N$-gram occupies $(100+1+1)$ KB, i.e., $0.1$ MB.

	\subsection{Model Loading}
	The storage strategy of distributing model over multiple files enables parallel loading. The VocabTrie is loaded to RAM that allows lookup by word or ID. For $N$-gram data, post decompression, all the elements are inserted in a vector-based structure ($O\left(1\right)$ insertion). The Prediction Optimisation block is loaded in a vector while the Completion Optimisation block is loaded into a map of vectors. The $N$-gram model is memory mapped to respective uni, bi and trigram vectors. The Class-$N$-gram word-to-class mapping is loaded into the same unigram vector. The other Class-$N$-gram blocks are loaded into similar structure. This helps in effectively utilising the run time memory.

	\subsection{Model Inference on mobile device}\label{sec:modelInference}
	The text suggestion in Op-Ngram involves Word Completion (WC) and Next Word Prediction (NWP). Word Completion refers to generating list of suggestions to complete current composing word. Next Word Prediction refers to suggesting prediction candidates for immediate next word, given a sequence of previous words. This subsection details the inference engine optimisations to maximise the relevance of top-$K$ suggestions.

	\subsubsection{Word Completion} WC suggestions are generated by performing lookup on the VocabTrie for words with matching prefixes. The obtained suggestion list is then sorted based on their probabilities of occurrence in the current context as per trigram and bigram information in $N$-gram model. The top-$K$ words in this sorted list are used as WC suggestions. Whereas if current composing word has single character, the word is picked from  Completion Optimisation block. Single character lookup in the trie is expected to yield many matching candidates, when only a limited number of same high-probability words would be selected for Word Completion suggestion. So, Completion Optimisation helps in reducing average WC building time by avoiding VocabTrie lookup.
	
	\subsubsection{Next Word Prediction} In general $N$-gram model, the conditional probability of a candidate word, $w_k$, is dependent on its immediate $\left(N-1\right)$ previous words \cite{brown1992class}. Mathematically, for $k \ge N$,
	
	\begin{equation}
		P\left(w_k\middle|w^{k-1}_{1}\right) = P\left(w_k\middle|w^{k-1}_{k-N+1}\right)
	\end{equation}
	
	In our trigram model ($N=3$), the probability of candidate word, given by $P\left(w_k\middle|w^{k-1}_{k-N+1}\right)$, is calculated as follows:
		
	\begin{equation}\label{eqn:inference}
	P\left(w_k\middle|w^{k-1}_{k-N+1}\right) =
		\begin{cases}
			\begin{aligned}[c]
				\frac{\left\lvert w^k_{k-2} \right\rvert}{\left\lvert w^{k-1}_{k-2} \right\rvert}, &&
				\left\lvert w^k_{k-2} \right\rvert > 0
			\end{aligned}\\
			\begin{aligned}[c]
				\frac{\left\lvert w^k_{k-1} \right\rvert}{\left\lvert w_{k-1} \right\rvert} \times \lambda, &&
				 \left\lvert w^k_{k-2} \right\rvert = 0\\
			
			\end{aligned}\\
			\begin{aligned}[c]

				\left(
					\begin{aligned}[c]
						r\cdot P'\left(w_k\middle|w^{k-1}_{k-N+1}\right)\\
						+
						\left(1-r\right)\cdot\left\lvert w_k \right\rvert
					\end{aligned}
				\right) \times \lambda^2,
			\end{aligned}\\
			\hfill{}
				\text{otherwise}
		\end{cases}	 
	\end{equation}
	
	where, $\left\lvert w^j_i \right\rvert$ represents the number of occurrences of the word sequence ($w_i$, $w_{i+1}$, ..., $w_j$), $\lambda$ represents a constant Stupid Backoff factor for falling back to the next lower order, $r$ is a ratio $\left(r\in[0,1]\right)$ and $P'\left(w_k\middle|w^{k-1}_{k-N+1}\right)$ denotes the class probability given by:
	
	\begin{equation}
		P'\left(w_k\middle|w^{k-1}_{k-N+1}\right) = 
		P\left(w_k\middle|C_k\right) \cdot P\left(C_k\middle|C^{k-1}_{k-N+1}\right)
	\end{equation}
	
\begin{figure}
		\centering
		\includegraphics[width=\linewidth]{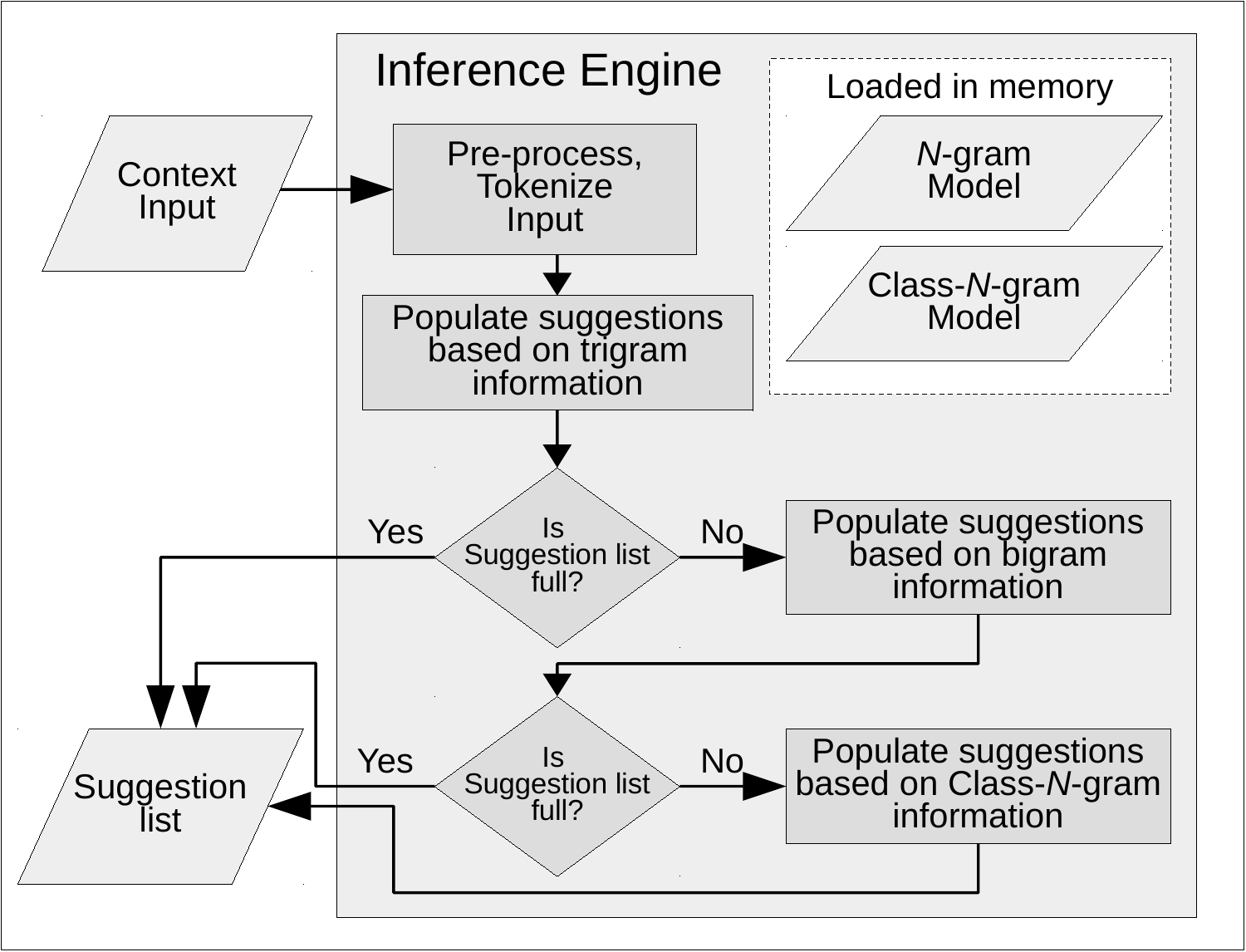}
		\caption{Inference Engine Flowchart for Next Word Prediction}
		\label{fig:nextWordPredictionFlowchart}
	\end{figure}

	Here, $P\left(w_k\middle|C_k\right)$ denotes the emission probability of word $w_k$ being mapped to class $C_k$, and $P\left(C_k\middle|C^{k-1}_{k-N+1}\right)$ indicates the transition probability of class $C_k$ following the class sequence ($C_{k-N+1}$, ..., $C_{k-1}$). The flowchart given in Figure \ref{fig:nextWordPredictionFlowchart} presents the suggestion building steps.
	
	In our inference engine, as in the case of corpus pre-processing phase, the current input sequence is prefixed with \texttt{<s>} tag (which becomes $w_1$), and NWP using $N$-gram and Class-$N$-gram information is performed only after the first input word has been entered, i.e, for words with $k \ge 3$. For $k = 2$ (first word), equation \ref{eqn:inference} is not applicable, and here, we leverage the list of bigrams with \texttt{<s>} as the first word (context). In cases, where top-$K$ suggestions cannot be obtained from both $N$-gram model and Class-$N$-gram model, Op-Ngram populates suggestion list via Prediction Optimisation of FWO block. This acts as a default suggestion list for a completely unseen context.

\begin{figure}
		\centering
		\includegraphics[width=0.97\linewidth]{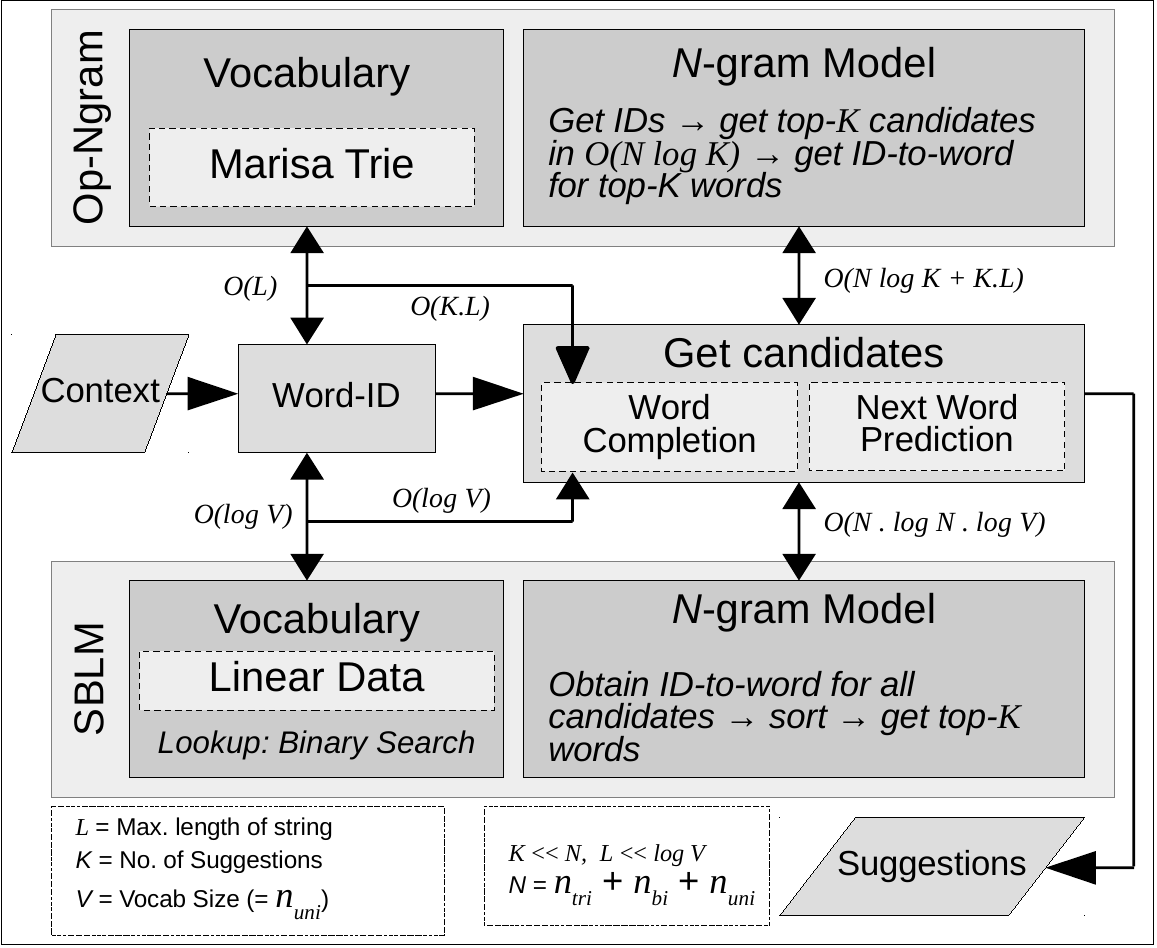}
		\caption{Algorithmic optimisation over SBLM}
		\label{fig:algorithmicOptimization}
	\end{figure}

	Op-Ngram algorithm has been highly optimized for extracting top-$K$ suggestions given context and prefix. We compare it in detail with the SBLM retrieval logic in Figure \ref{fig:algorithmicOptimization}. In Op-Ngram, given the context, the context ID is obtained from VocabTrie. The top-$K$ model inference for this context is obtained from the $N$-gram and Class-$N$-gram. The resultant IDs are used to retrieve the suggestion words by VocabTrie lookup.
	
	Time complexity for prediction is $O(N \log K + KL)$ (SBLM: $O(N \log N \cdot \log n_{uni})$), and for completion is $O(CL + C \log(n_{tri} \cdot n_{bi}) + N \log K)$  (SBLM: $O(C \log(n_{tri} \cdot n_{bi} \cdot n_{uni}) + N \log N)$). Here, $C$ denotes the number of candidates given the prefix (typed partial word). (Refer legend of Figure \ref{fig:algorithmicOptimization} for remaining notations).
	
	\section{Experiment Results}\label{sec:experimentResults}
	
	We evaluate Op-Ngram and present all the comparison results for English and a morpheme language Hindi. The KSR and NWP metrics have also been evaluated for multiple languages. For English, datasets from GMB 2.2.0 \cite{dataset_gmb220} and CoNLL 2000 \cite{dataset_conll2000} have been sampled, each containing $\sim 10 k$ words. For Hindi and other languages, due to lack of widely used datasets, $\sim 10k$-word samples from linguist-verified testsets have been used. Statistics for these samples have been laid out in Table \ref{tab:testsetDescription}. For evaluation purpose, we have used a Samsung Galaxy S8 device.
	
	\begin{table}[h]	
		\small
		\centering
		\resizebox{\columnwidth}{!}{%
			\begin{tabular}{c c c c c c}
				\toprule
				Alias & Language & Source & Lines  & Words   & Characters \\  
					\midrule
				 EN1  & English  & GMB 2.2.0  & $650$  & $10301$ &$60134$\\
				 EN2  & English  & CoNLL 2000 & $740$  & $10250$ &$58347$\\
				 HI   & Hindi    & Linguist   & $1291$ & $13018$ &$153994$\\
				 KN   & Kannada  & Linguist   & $1398$ & $12603$ &$226227$\\
				 MR   & Marathi  & Linguist   & $1374$ & $15616$ &$250783$\\
				 TA   & Tamil    & Linguist   & $1368$ & $11128$ &$211822$\\
				 TE   & Telugu   & Linguist   & $1390$ & $14622$ &$276847$\\ \bottomrule
			\end{tabular}%
		}
		\caption{Description of testset samples}\label{tab:testsetDescription}
	\end{table}
	
	\paragraph{Model Storage} Table \ref{tab:romSizeComparison} shows that our ROM size is less than both SBLM and KenLM. The reasons are (a) our model stores vocabulary words in a VocabTrie, (b) $N$-gram data is stored with reference to $N$-gram IDs,  (c) the effective number of classes used in Op-Ngram is $\sim32$, (d) the most likely class block is stored as a sequence indexed by implicit context.
	\begin{table}[ht]
		\small
		\centering
		\begin{tabular}{c c c}
			\toprule
			& Vocab  + $N$-gram + Class-$N$-gram & Total           \\ \midrule
			SBLM  & $3.50$                             & $3.50$          \\
			KenLM    & $5.12$                             & $5.12$          \\
			Op-Ngram & $0.80$ + $1.40$ + $0.10$           & $\textbf{2.30}$ \\ \bottomrule
		\end{tabular}
		\caption{Compressed ROM size (in MB) comparison}\label{tab:romSizeComparison}
	\end{table}

	\paragraph{Loading Time and RAM Size} The loading time is measured as the time taken to load the LM resources to memory that also includes the time taken to decompress the compressed $N$-gram data file in RAM. The improvement in loading time of the compressed model outweighs the additional decompression overhead. The loading time is further reduced by parallel loading of the VocabTrie, $N$-gram data and Class-$N$-gram files. Table \ref{tab:loadingTimeAndramSizeComparison} compares Op-Ngram loading time with that of SBLM.
	
	\begin{table}[ht]
		\small
		\centering
		\begin{tabular}{c r c c}
			\toprule
			               & SBLM     & Op-Ngram & Improvement (\%) \\ \midrule
			Load Time (ms) & $549.30$ & $66.30$  & $\textbf{87.93}$ \\
			RAM Size (MB)  & $44.00$  & $10.47$  & $\textbf{76.20}$ \\ \bottomrule
		\end{tabular}
		\caption{Loading Time and RAM size comparison on mobile device}\label{tab:loadingTimeAndramSizeComparison}
	\end{table}

	\paragraph{Average Suggestion Time} Average Suggestion Time refers to the weighted mean of time taken for WC and NWP. Table \ref{tab:retrievalTimeComparison} shows upto 89\% improvement with Op-Ngram in terms of retrieval time.
	
	\begin{table}[ht]
		\small
		\centering
		\begin{tabular}{c c S[table-format=3.2] c c}
			\toprule
			\multirow{2}{*}{Language} & \multirow{2}{*}{Testset} & \multicolumn{1}{c}{SBLM} & \multicolumn{1}{c}{Op-Ngram} & \multicolumn{1}{c}{Improvement} \\
			                          &                          & \multicolumn{1}{c}{(ms)} & \multicolumn{1}{c}{(ms)}     & \multicolumn{1}{c}{(\%)}        \\ \midrule
			         English          & EN1                      & 15.16                    & 2.40                         & \textbf{84.17}                  \\
			         English          & EN2                      & 16.08                    & 2.40                         & \textbf{85.07}                  \\
			          Hindi           & HI                       & 20.50                    & 2.20                         & \textbf{89.27}                  \\ \bottomrule
		\end{tabular}
		\caption{Comparison of Average Suggestion Time on mobile device}
		\label{tab:retrievalTimeComparison}
	\end{table}

	\subsubsection{KSR and NWP} Keystroke Saving Ratio ($M_{KSR}$) and Next Word Prediction ($M_{NWP}$) are two important metrics used to evaluate LM. KSR indicates the number of keystrokes saved due to effective suggestions provided by the engine. It is defined as:
	
	\begin{equation}
		M_{KSR} = \frac{n_c - n_k}{n_c} \times 100\%
	\end{equation}
	
	where $n_c$ denotes the number of actual characters in test set and $n_k$ denotes the number of keystrokes which were needed to produce the test set in soft keyboard\cite{Trnka:2007:CSW:1296843.1296877}. NWP refers to the probability that given a sequence of words from a sentence in test data, $w_1^{k-1}$, the next word in that sentence, $w_k$ will appear among the word prediction candidates. Mathematically,
	
	\begin{equation}
		\small
		M_{NWP} = \frac{\mathlarger{\sum}_{\substack{s \in \textbf{S},\\ k \in [1, n^s_w]}}\left\lvert \left\{ w^s_k \middle| w^s_k \in \textbf{W}^s_k\right\}\right\rvert}{\mathlarger{\sum}\limits_{s \in \textbf{S}} n^s_w} \times 100\%
	\end{equation}
	
	where $\textbf{S}$ refers to the set of all sentences in test set, $w^s_k$ refers to actual test set word at $k^\text{th}$ position in sentence $s$, $\textbf{W}^s_k$ denote the set of suggestions provided by the inference engine as candidates for word $w^s_k$, and $n^s_w$ indicate the total number of words in sentence $s$ of test set.
		
	Table \ref{tab:enUsKsrNwp} compares the KSR and NWP for English, which shows that in spite of the smaller model size of Op-Ngram, its performance is as good as SBLM and KenLM.
	Table \ref{tab:tymeNgramVsXt9KsrNwp} compares Op-Ngram with KenLM and demonstrates that our model outperforms KenLM in giving relevant predictions for multiple languages.
	
		\begin{table}[ht]
		\small
		\centering
		\begin{tabular}{c c c c c c c}
			\toprule
			\multirow{3}{*}{Testset} &       \multicolumn{3}{c}{KSR (\%)}        &       \multicolumn{3}{c}{NWP (\%)}        \\
			     \cmidrule{2-7}      & \multirow{2}{*}{SBLM} & Ken-    & Op-     & \multirow{2}{*}{SBLM} & Ken-    & Op-     \\
			                         &                       & LM      & Ngram   &                       & LM      & Ngram   \\ \midrule
			          EN1            & $55.51$               & $57.64$ & $57.01$ & $20.82$               & $22.96$ & $22.26$ \\
			          EN2            & $53.17$               & $53.89$ & $53.19$ & $18.97$               & $21.19$ & $20.31$ \\ \bottomrule
		\end{tabular}
		\caption{KPI comparison between SBLM, KenLM and Op-Ngram for English}\label{tab:enUsKsrNwp}
	\end{table}
	
	\begin{table}[ht]
		\small
		\centering
		\resizebox{\columnwidth}{!}{%
			\begin{tabular}{c c c c c c}
				\toprule
				\multirow{2}{*}{Language} & \multirow{2}{*}{Testset} & \multicolumn{2}{c}{KSR (\%)} & \multicolumn{2}{c}{NWP (\%)} \\
				     \cmidrule{3-6}       &                          & KenLM   & Op-Ngram           & KenLM   & Op-Ngram           \\ \midrule
				          Hindi           & HI                       & $48.20$ & $50.12$            & $19.58$ & $22.25$            \\
				         Kannada          & KN                       & $31.34$ & $41.77$            & $4.09$  & $9.67$             \\
				         Marathi          & MR                       & $33.32$ & $41.62$            & $7.54$  & $15.88$            \\
				          Tamil           & TA                       & $39.04$ & $45.98$            & $4.72$  & $10.33$            \\
				         Telugu           & TE                       & $40.93$ & $47.64$            & $5.68$  & $12.30$            \\ \bottomrule
			\end{tabular}%
		}
		\caption{Op-Ngram outperforms KenLM for multiple Indian languages}\label{tab:tymeNgramVsXt9KsrNwp}
	\end{table}
	
	\section{Conclusion}
	Our contributions are two-fold -- (a) a light-weight model, and (b) a novel storage strategy that is specialised for the speed and memory. Our technique outperforms SBLM by a large margin in various aspects, while maintaining similar KSR and NWP.
	
	\section{Future Work}
	Currently, the model inference is based on a single language model generated from a single corpus. We seek to explore the possibility of a Hybrid $N$-gram model that merges various domain specific pre-trained model into a single model. We expect this approach to help in generating a light-weight model with improved text suggestion. The text suggestion can also be improvised using on-device learning to work in tandem with Op-Ngram Model. On-device learning can be achieved by generating a dynamic $N$-gram model based on the history of user context. The interpolation of pre-trained and dynamic $N$-gram models becomes a crucial problem to be explored. 
	
	\bibliographystyle{IEEEtran}
	\bibliography{bibliography}\nocite{*}
	
\end{document}